\documentclass[final,3p,times,twocolumn]{elsarticle}

\usepackage{hyperref}
\usepackage{natbib}
\usepackage{url}
\usepackage{xcolor}
\usepackage{numcompress}
\usepackage{amsmath}
\usepackage{multirow}
\usepackage{amssymb}
\usepackage{pifont}
\usepackage{adjustbox,lipsum}

\newcommand{\cmark}{\ding{51}}%
\newcommand{\xmark}{\ding{55}}%
\usepackage{algpseudocode,algorithm,algorithmicx}
\usepackage{caption}
\makeatletter
\algrenewcommand\ALG@beginalgorithmic{\tiny}
\makeatother
\newcommand*\Let[2]{\State #1 $\gets$ #2}
\algblock{inputs}{endinputs}
\algnotext{endinputs}
\algblock{local}{endlocal}
\algnotext{endlocal}
\algblock{outputs}{endoutputs}
\algnotext{endoutputs}
\algblock{require}{endrequire}
\algnotext{endrequire}
\newcommand{\Desc}[2]{\State \makebox[2em][l]{#1}\enspace\enspace{#2}}

\journal{Applied Soft Computing}

\makeatletter \def\ps@pprintTitle{  \let\@oddhead\@empty  \let\@evenhead\@empty  
\def\@oddfoot{\hfill\thepage}  \def\@evenfoot{\thepage\hfill}} \makeatother

\begin{document}

\begin{frontmatter}

\title{Comparing seven methods for state-of-health time series prediction for the lithium-ion battery packs of forklifts\tnoteref{SoHmodels}}

\author[1]{Matti Huotari}
\author[2]{Shashank Arora}
\author[1,4]{Avleen Malhi}
\author[1,3]{Kary Fr{\"a}mling}
\address[1]{Department of Computer Science, Aalto University, Espoo, Finland.}
\address[2]{Department of Mechanical Engineering, Aalto University, Espoo, Finland.}
\address[3]{Department of Computer Science, Ume\r{a} University, Ume\r{a}, Sweden.}
\address[4]{Department of Computing and Informatics, Bournemouth University, Bournemouth, UK.}

\begin{abstract}

A key aspect for the forklifts is the state-of-health (SoH) assessment to ensure the safety and the reliability of uninterrupted power source. Forecasting the battery SoH well is imperative to enable preventive maintenance and hence to reduce the costs. This paper demonstrates the capabilities of gradient boosting regression for predicting the SoH timeseries under circumstances when there is little prior information available about the batteries. We compared the gradient boosting method with light gradient boosting, extra trees, extreme gradient boosting, random forests, long short-term memory networks and with combined  convolutional neural network and long short-term memory networks methods. We used multiple predictors and lagged target signal decomposition results as additional predictors and compared the yielded prediction results with different sets of predictors for each method. For this work, we are in possession of a unique data set of 45 lithium-ion battery packs with large variation in the data. The best model that we derived was validated by a novel walk-forward algorithm that also calculates  point-wise confidence intervals for the predictions; we yielded reasonable predictions and confidence intervals for the predictions. Furthermore, we verified this model against five other lithium-ion battery packs; the best model generalised to greater extent to this set of battery packs. The results about the final model suggest that we were able to enhance the results in respect to previously developed models. Moreover, we further validated the model for extracting cycle counts presented in our previous work with data from new forklifts; their battery packs completed around 3000 cycles in a 10-year service period, which corresponds to the cycle life for commercial Nickel-Cobalt-Manganese (NMC) cells.

\end{abstract}


\end{frontmatter}


\section{Introduction} \label{intoduction}

Efficient transportation systems can improve the flow of goods and diminish the amount of energy used. In this regard, electric vehicles (EVs) have gained attention, especially in the area of forklifts, that now are not only energy-efficient but also safer for the drivers as they produce no fumes, vibrate less and are quieter than  combustion-engine-powered forklift trucks. Lithium-ion batteries are widely employed to power these EVs. A battery's ability to store and deliver electrical energy is expressed by a measure, battery state of health (SoH), that is used for monitoring and for controlling these batteries in order to maximize their availability for operation. Heat generation in battery packs increases as they age. If unchecked, it can cause internal short circuits and compromise the safety of EV passengers and first responders. Therefore, battery packs in EVs are replaced when their SoH decreases below 80\% \cite{jal15cyc}. However, these lithium-ion batteries can still be used for less-demanding grid-connected energy storage applications, i.e., the batteries can have a second-life use as components of an energy storing system for the sustainable energy management of a Smart City \cite{priyadarshini2020new, martinez2016evaluation, calvillo2016energy}.

Overall, making battery related forecasts remains difficult in such manner that it generalizes well \cite{nuh13hea}. For data-based approaches, there is a trade-off between complex hypotheses that fit the training data well, and simpler hypotheses that may generalize better \cite{rus13AI, box08tim}. For this paper, we are in possession of a unique data set of 45 lithium-ion battery packs with large variation in the data \cite{huotari2020git}. We compared five regression methods for making predictions for the SoH timeseries:  gradient boosting (GB) \cite{has17elewf}, light gradient boosting (LGB) \cite{ke2017lgbm},  random forests (RF)\cite{has17elewf}, randomised extra trees (ETR) \cite{geurts2006etr}, extreme gradient boosting (XGB) \cite{chen2015xgboost}. We also compared two other methods: long short-term memory networks (LSTM) \cite{hochreiter1997long} and long short-term memory network and convolutional neural network (LSTM-CNN) method \cite{zhao2017convolutional}. Based on these methods, several models were grid searched to yield the best hyperparameters and the best model for each of the methods. Then the models yielded were back-tested in walk-forward manner \cite{schumann2018backtesting}. We  created a novel walk-forward algorithm for the data set that was re-framed as a supervised learning problem, that can utilize less steps and is therefore computationally lighter than the one-step-ahead walk-forward algorithm \cite{MATLAB2021, uber2021slideexpand}; furthermore, the novel algorithm can be utilized to calculate the point-wise confidence intervals for the predictions. With this  novel algorithm, we examined several prediction-period related parameters to detect the best time-span to utilize in order to yield reliable predictions. Moreover, as there is data available from several battery packs, we assessed the applicability of this data to the finally yielded best model, and then we verified the final model against five similar lithium-ion battery packs.

In the selection of the machine learning methods, we wanted to use methods that adapt well to timeseries regression, which, e.g., do not shuffle or split the data in such manner that the timeseries sequence is broken. Secondly, we aimed at using a method with a fairly good record of speed and model performance in the field of machine learning, such as XGB \cite{chen2015xgboost}. Thirdly, we applied empirical mode decomposition to the SoH signal to determine if the decomposition data enhances the model performance \cite{luukko2016introducing}. Little research exists related to battery packs' state-of-health predictions; therefore, we have identified a few research gaps, based on which we have formulated the following research questions: \begin{enumerate}
    \item How can we obtain a robust SoH estimation model for lithium-ion battery packs with low error margins?
    \item How can we validate the correctness of the used model?
\end{enumerate} 

\subsection*{Contributions}

This paper contributes to the literature by introducing a novel GB model for SoH prediction based on real-world application of lithium-ion battery packs in forklifts and implementation of a novel walk-forward algorithm  \cite{has17elewf,diciccio1996bootstrap} for validating the models. The implementation of the novel walk-forward algorithm was first tested against a public data set on household power consumption \cite{Hebrail2010}. Moreover, we further validated the model for extracting cycle counts presented in our previous work \cite{huo20dyn} with data from new forklifts; their battery packs completed around 3000 cycles in a 10-year service period, which corresponds to the cycle life for commercial Nickel-Cobalt-Manganese (NMC) cells published in \cite{jal15cyc}. 

\subsection*{Organization}

The remainder of the paper proceeds as follows. First, the related literature in presented section 2 and the materials and methods in section 3. The model development steps are described in the section 4. Subsequently, the main findings are presented in section 5, followed by the discussion in section 6. The conclusions and future work are discussed in section \ref{conclusion}. 

\section{Literature Review}
The lithium-ion battery packs in EVs and grid storage systems can benefit from the added reliability and safety assurance provided by a fast, yet accurate, SoH prediction. Traditionally, SoH forecasting has relied on equivalent-circuit models; however, more recently statistical and machine-learning techniques have been proposed, including ARIMA based statistical methods \cite{koz03ele, mak00M3, mak18M4}, neural networks \cite{koz03ele, han18NN, andre2013comparative}, Gaussian processes \cite{ric17gau, zha20ide}, support vector machines \cite{sah08unc, lei20mul, benkedjouh2013remaining}, ensemble machine learning methods \cite{sev19dat, att20clo} and the XGB \cite{koz03ele, SEPASI2014368}. The success of the studies cited above demonstrates the capabilities of these approaches. Nevertheless, it is also known that the relationship between the basic signals and the SoH is complex under real conditions \cite{A123_14bat, zha20per}. In addition, these studies focus on single cells, although the voltages available through the state-of-the-art single cells are insufficient for supporting an electric driveline. Therefore, many cells needs to be combined in series and in parallel to build up battery packs that are then used as energy sources in the EVs. Due to the manufacturing inconsistencies and differences in working environments, the behavior of each battery pack varies in real-life applications. Furthermore, this variation will become even larger as the cells age. As a result, an estimation based on the unit cell model will be inaccurate for real-world applications \cite{hua2015soh}.

Therefore, in this study, in contrast to several recent studies that model a battery pack based on unit cells (e.g., \cite{ren2019soh, zhang2018soh, wang2016soh}), we employed a battery data set from lithium-ion battery packs that are used in electric forklifts \cite{huotari2020git}. The data was obtained from one of the companies in the industry. This data is unique to our knowledge, as we have not found data repositories related to the forklift battery data, except on the cell level \cite{saha2007battery}. 

\section{Materials and methods} \label{materials}
\subsection{Data set description}

For this study, the data set consists of 45 three-year timeseries that are derived from lithium-ion battery packs used in electric forklifts. The nominal capacity of the battery packs is 220 kW. We were told that the data was collected from different countries and continents. From the data we observed that the mean monthly ambient temperature for the forklifts ranged from 21.5$^{\circ}$C to 32.3$^{\circ}$C with instantaneous temperatures beyond this range. The data was collected using sensors selected by the battery manufacturer that were attached to the batteries. The raw data was sent to a local hub according to the date of the data collection and the serial number of the battery. The selected data points, comprised of current, voltage, and the ambient temperature, were taken every minute (Table \ref{tabel:basicunits}). Based on these data points, we extracted more features using our feature extraction method  \cite{huo20dyn}. 

\begin{table}[hbt!]
\centering
\scriptsize
\caption{Summary of the basic signals.}
\begin{tabular}{|l|}
\hline
\textbf{Basic signals}                         \\ \hline
Time stamp of the data, 1 min interval \\ \hline
Measured voltage V                    \\ \hline
Measured current A                    \\ \hline
SOC \%                                \\ \hline
Ambient temperature $^{\circ}$C         \\ \hline
\end{tabular}
\label{tabel:basicunits}
\end{table}

In this study, the number of timeseries available was greater than in our previous study \cite{huo20dyn}; however, the new data supported our previous findings. We observed that the median number of occurrences of the charging pulses was two in a day; here a charging pulse is defined as a period between 5\textendash30 minutes when the battery's state-of-charge increases.  However, there were days when there were no charging pulses (timeseries are irregular). For more information on the basic and on the derived signals, see our previous paper \cite{huo20dyn}. 

\subsection{Proposed Methodology} \label{methodology}

Although there are several statistical methods \cite{pratama2016missingvalues} or neural network based methods \cite{che2018recurrent} to compose a solution for incomplete or irregular timeseries, in this paper we wanted to utilise regression methods that require regular timeseries. Furthermore, we wanted to compare the results with our previous ARIMA based results with results that are directly comparable \cite{huo20dyn}. For these reasons, the irregular timeseries based on one-minute measurements was aggregated to regular daily timeseries; the number of resulting timesteps was around 1000 for the battery packs. 

The overall prediction target, state-of-health, is defined as the ratio of current capacity to the initial capacity of a battery; the SoH is denoted as timeseries in this paper \cite{huo20dyn}:  

\begin{equation}\label{eq:1}
SoH(t) = \frac{C_n(t)}{C_0(t_0)}     
\end{equation}

The empirical mode decomposition is a mathematical time domain decomposition method, which can convert a group of timeseries into locally narrow band components, the intrinsic mode functions \cite{huang1998empirical}. This method is applied to, e.g., asserting power quality \cite{babu2017fault}, or predicting remaining useful lifetime of lithium-ion batteries \cite{koz03ele}.

A timeseries can be transformed by using an empirical mode decomposition, which in this case is denoted as: 
\begin{equation}\label{eq:IMF}
SoH(t)=\sum_{i=0}^{N}c_{i}(t)+r_{N}(t)
\end{equation}

where $c_{i}(t)$ are the intrinsic mode functions (IMFs) separated by instantaneous frequencies, $r_{N}(t)$ is the residue and $N$ is the finite amount of decompositions obtained \cite{huang1998empirical}. In this paper, for the needs of the model, the residue is used as the trend \cite{Wu14889}, although there are more refined trend extraction methods available \cite{6292713}.

The IMFs can be transformed by using the Huang-Hilbert transform. This is used to obtain the analytic signal, which can be presented in polar form \cite{RAWLINS2000427}, neglecting the residue. In this case this yields:  

\begin{equation}\label{eq:if1}
SoH(t) = \sum_{j = 1}^{n} a_{j} (t) e^{i \left(\theta\right)_{j} (t)}
\end{equation}

where $a_{j}(t)$ are the analytic signals \cite{BACCIGALUPI20161}. The synthesised signal models a non-stationary and non-linear system analytically, i.e.,
\begin{equation}\label{eq:analytic}
SoH(t) = a(t) cos\theta(t) 
\end{equation}

where where $a(t)$ is the instantaneous amplitude and $\theta(t)$ the instantaneous phase. The optimal values for the parameters of the synthesised signal models are tuned according to the signal in question; i.e, the IMFs and residue change as the data set changes, and decomposing parameters needs to be set accordingly to yield an accurate decomposition \cite{luukko2016introducing}. 
Finally, the instantaneous frequency is obtained through the derivative of the instantaneous phase, i.e.,

\begin{equation}\label{eq:ifreq}
f(t) = \frac{1}{2 \pi} \frac{d \theta (t)}{dt} 
\end{equation}

The derived $f(t)$ is a tool for analyzing transient signals, such as battery pack SoH, whose constituent frequencies may change over time \cite{boashash1992estimating,yan2006hhhealth}. The lagged instantaneous frequency, $f(t-1)$, was used as one of the predictors for our models. 

\subsection{Selected SoH prediction methods} \label{regression}

Extreme gradient boosting (XGB) is an ensemble of gradient boosted decision trees algorithm \cite{has17elewf,che16XGB}. It uses decision trees where new trees improve the model consisting of those trees that are already part of the model. It is used, for example, for forecasting energy load  \cite{yucong2020xgbc} and for forecasting the battery cell state-of-charge (SoC) as represented in \cite{DINEVA2021102351}. As  for the other methods  utilized,  gradient boosting (GB) is described in \cite{has17elewf}, extremely randomised trees (ETR) in \cite{geurts2006etr}, random forests (RF) in \cite{has17elewf}, long short-term memory networks (LSTM) e.g. in \cite{ke2017lgbm} and convolutional neural networks (CNN) in \cite{zhao2017convolutional}.

\subsection{Data preprocessing methods and  performance metrics}

The outliers in the initial data for each battery pack were eliminated by the interquartile range (IQR) method \cite{zwillinger1999crc}.  After removing the outliers, we imputed some missing daily data. The missing values were the mean of the daily values above and below the missing values.

After creating the predictors and targets for the models, we re-framed the multivariate timeseries as a supervised learning problem \cite{has17elewf} in order to define the number of past time steps used for making a forecast and to define the number of prediction timesteps for the prediction horizon.
In the model tuning phase, we split this re-framed data to training data and test data (more details on splitting methods and on numerical values utilised for the test and train sets is in the next section in Table \ref{table:modelcomp}). We evaluated the used prediction methods using four different metrics: the root-mean-squared error (RMSE) \cite{RMSE}, the mean absolute error (MAE) \cite{MAE}, coefficient of determination, R$^2$ \cite{R2} and the explained variance (EVAR) as in Equation \ref{eq:13}. In this paper, the MAE and the RMSE are related to the SoH range 0-100(+) \%. E.g., a MAE 1 implies that, on average, the forecast's distance from the true SoH value is 1. For this data set, a MAE value of 1 is significant as, e.g., the yearly degradation of SoH for battery (a) is around 2.2 percentage points \cite{huo20dyn}.

\begin{equation}\label{eq:13}
\text{EVAR} = 1 - \frac{Var\{ y - \hat{y}\}}{Var\{y\}}
\end{equation}

For EVAR and R$^2$ evaluation methods, the best possible score is 1.0; a baseline model predicting the  mean ($\bar{y}$) has score 0; models with less skill than the baseline model will have negative scores.

\subsection{Validation of models and calculation of the point-wise confidence intervals for the estimates by a novel walk-forward algorithm} \label{walk}

We executed a basic comparison of models with walk-forward method to predict and to find the best model in terms of MAE \cite{schumann2018backtesting, MATLAB2021}. A basic walk-forward method, that utilises expanding window and proceeds one-step-ahead at each iteration round, can be utilised for all comparisons \cite{uber2021slideexpand}. Nevertheless, we wanted to find an approach that utilise computational resources sparingly as the timeseries can grow long (10-years or more) or there can be several models to be evaluated at the same time for a fleet of fork-lifts. The following parameters can be set for the novel algorithm: sample size (number of the latest observation windows used by the algorithm) and roll size (number of windows stepped over in an iteration) \ref{alg:wf-pci}; these functionalities are not part of the standard machine learning (sklearn) library timeseries split for a multivariate data set re-framed as supervised learning problem in a simple manner. For the implementation of the algorithm, we used Pandas's append and del functions and sklearn's regression methods (GB, RF, XGB and ETR) \cite{reback2020pandas}. It is noteworthy that the window size (number of past observations and future observation in the prediction window) was set in the model tuning phase; furthermore, each window size requires a model of its own \cite{simpson2009math}. 

\begin{algorithm}[hbt!]
  \captionsetup{font=scriptsize}
  \centering
  \caption{Walk-forward: sample predictions with point-wise confidence intervals}
  \label{alg:wf-pci}
  \begin{algorithmic}[1]
     
    \inputs\textbf{:}
    \Desc{$Obs$:}{timeseries' observations that are re-framed as supervised learning problem (re-framed to windows)}
    \Desc{$n_{S}$:}{sample of windows utilised for testing}
    \Desc{$n_{roll}$:}{number of windows stepped over in a window roll}
    \endinputs
    
    \local\textbf{ variables:}
    \Desc{$Te$:}{a sample of timeseries' windows}
    \Desc{$Tr$:}{timeseries' windows preceding $Te$}
    \Desc{$RTe$:}{a rolled sample of timeseries' windows}
    \Desc{$W_{ {p}}$:}{consequent predictors utilised for making predictions}
    \endlocal
    
    \outputs\textbf{:}
    \Desc{$\widehat{SoH}$:}{SoH predictions for the sample}
    \Desc{$CI$:}{point-wise confidence interval for these SoH predictions}
   \Desc{$W_{ {t}}$:}{consequent targets (ground truth)}
    \endoutputs
    
    \require\textbf{:}{}
    \Desc{$|Obs| > 1$}{}
    \Desc{$n_{S} > 0$}{} 
    \Desc{}{}
    \endrequire

\Let{$Tr$,$Te$}{split $Obs$ to train and test sets according to $n_{S}$}
\Let{$RTe$}{[\enspace]}
\For{$R \gets 0 \textrm{ to } |Te| \enspace \mathbf{mod} \enspace n_{roll}$}
    \Let{$RTe$}{append window $(Te[R])$}
\EndFor

\Let{$\widehat{SoH}$,$CI$}{[\enspace],\enspace[\enspace]}
    \For{$T \gets 0 \textrm{ to }|RTe|$}
   
   \Let{$W_{p},W_{t}$}{$RTe[T]$ separate predictors and targets for this iteration step}
   
   \Let{$\widehat{SoH}$}{fit the model with $Tr$ and predict with  $W_{p}$}
   
   \Let{$CI$}{append $\pm$1.98*$\mathbf{SE}$($\widehat{SoH}$)}\Comment{Eq. \ref{eq:15}}
   
   \Let{$Tr$}{append windows from $Te$ until and including window $RTe[T]$}
   \If{sliding window}
   \Let{$Tr$}{delete $|n_{roll}|$ windows from head of $Tr$}    
   \EndIf
\EndFor
\State \textbf{return} $\widehat{SoH}$, $W_t$, $CI$\
\end{algorithmic}
\end{algorithm}

As the algorithm utilises the sliding window method, the successive training sets are not super-sets of those coming before them, and this yields models that have more variation; however, at the same time, some of the training data is lost.

Moreover, the algorithm yields point-wise confidence intervals (CI) that quantify the uncertainty for the predictions \cite{MATLAB2021, zivot2006vector}. In this paper, we added upper and lower confidence intervals (CI) to each of the point-wise predictions for the selected final model (Figure \ref{fig:ci12}). The standard error (SE\textsubscript{SoH}) that is needed for yielding a confidence interval was calculated as:

\begin{equation}\label{eq:14}
\widehat{SE\textsubscript{SoH}}(\hat{\mu}_t(n))=\frac{\hat{\sigma}_t(n)}{\sqrt{n}}			\end{equation}
where $n$ are the prediction made by the novel Algorithm  \ref{alg:wf-pci}.

We used the 95\% significance level for the point-wise confidence intervals, which corresponds to the Gaussian distribution critical value 1.96. Hence, a confidence interval was calculated as:

\begin{equation}\label{eq:15}
CI = 1.96 \widehat{SE\textsubscript{SoH}}(\hat{\mu}_t(n))
\end{equation}
for a point-wise SoH prediction  (Algorithm \ref{alg:wf-pci} above)  \cite{MATLAB2021, zivot2006vector}.

\section{Development of models}

\begin{figure*}[hbt!]
\includegraphics[width=1.0\linewidth]{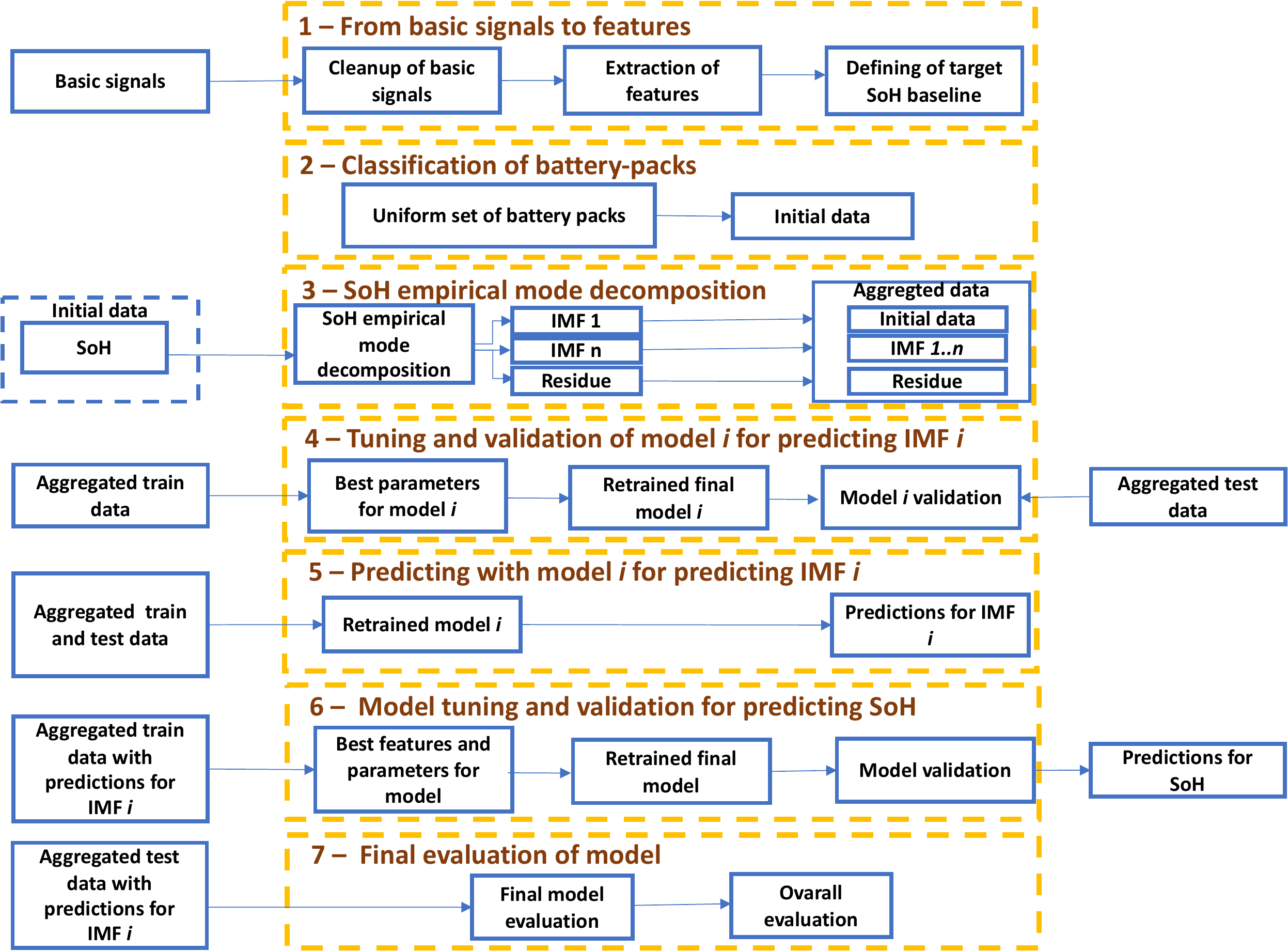}
\caption{The selected SoH prediction model development steps 1--2: data preparation. Steps 3--5: SoH decomposition and decomposition related predictions. Steps 6--7: the final model predicting the SoH with the model evaluation.}
\label{fig:data_model}
\end{figure*}

\begin{figure*}[hbt!]
\centering
\includegraphics[width=0.6\linewidth]{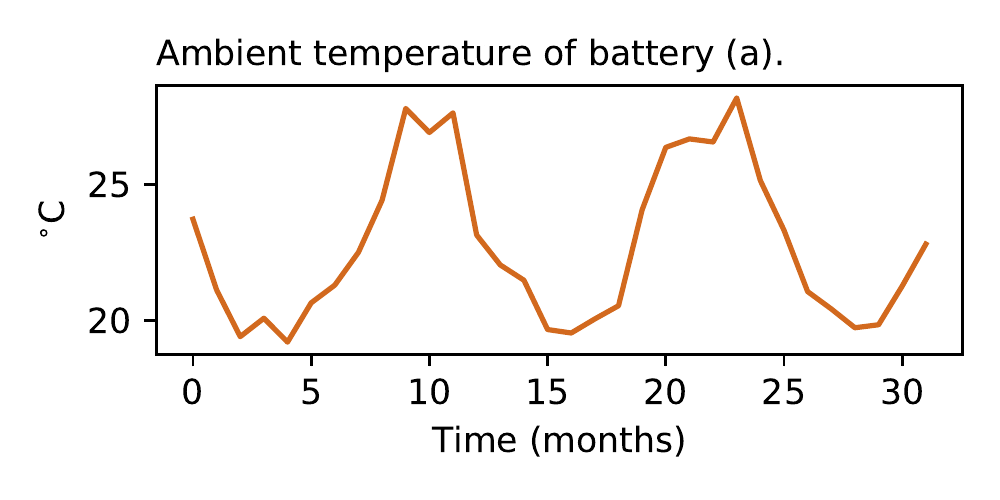}
\caption{Ambient temperature for battery (a)}
\label{fig:temp_A}
\end{figure*}

The overall flow of the model development is depicted in Figure \ref{fig:data_model}. In the steps 1\textendash 2, we cleaned the basic signals and extracted new ones \cite{huo20dyn}. Initially, we had to eliminate the adverse effect of severe transient failures, where, e.g, a sensor had sent erratic values.  Furthermore, for selecting the battery packs for the model development, we scrutinised the ambient temperatures. On one hand, ambient temperature below zero may have had an adverse effect on the SoH of the battery \cite{fle16int}. On the other hand, relatively high ambient temperatures ($>32 ^{\circ}$C) also have an adverse effect on the SoH \cite{jal15cyc, aro18sel, aro18ano}. Moreover, for the batteries used in forklifts, the ambient temperature fluctuation showed some seasonality for all battery packs; an example of this is in Figure \ref{fig:temp_A}. For these reasons, for the model development, we used a set of battery packs with the same 32-month data record and the mean ambient temperature between the range mentioned above. 

For the remaining battery packs, the following physical quantities were extracted: charging time, charging voltage, charging current and difference in state-of-charge (SOC) during chargings. From these we derived charging energy and charging cycles as described in \cite{huo20dyn}. 

\begin{figure*}[hbt!]
\centering
\includegraphics[width=1.0\linewidth]{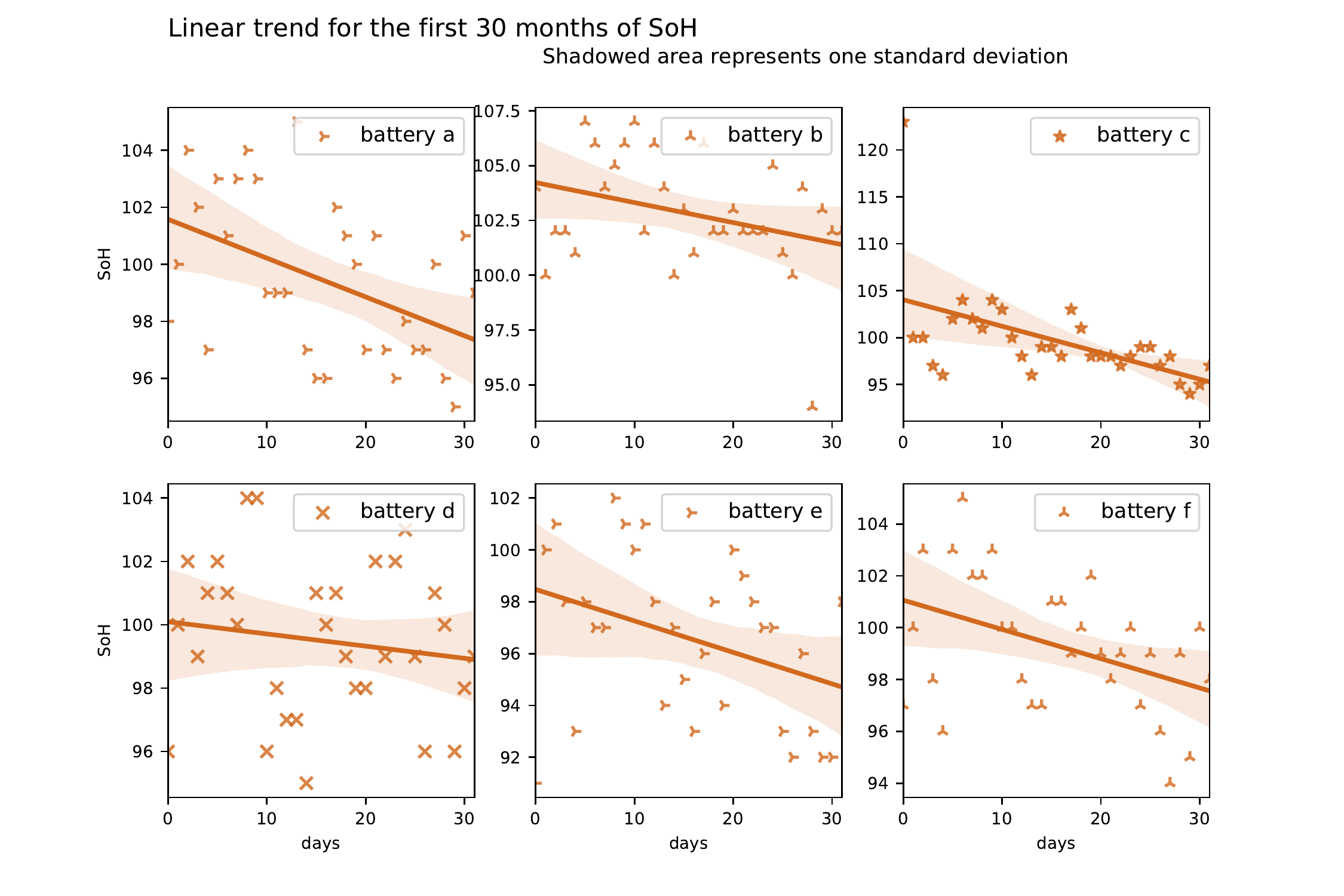}
\caption{State-of-health trends for batteries (a) -- (f).}
\label{fig:sohtrend}
\end{figure*}

For verifying that the  cycle count calculation method developed in the previous paper was valid for the new battery packs available for the study this time, we selected batteries’ (a\textendash f) timeseries randomly from the suite of 45 batteries for making graphs and for making initial reasoning based on those graphs; these new results supported the findings of our previous study. In the monthly averaged SoH data, there were some obvious outliers; moreover, there was remarkable fluctuation day-to-day. Overall, the SoH values for a typical battery showed a linearly fitted trend downwards. However, small increases in the capacity after a slow cycle or a rest period may result in the SoH exceeding 100\% \cite{sev19dat}. This can be visually confirmed from the plots for the selected six batteries in Figure \ref{fig:sohtrend}. We decided to use the battery (a) in the further development of the model in more detail; however all of the six batteries are taken into account in the final verification.

\begin{figure*}[hbt!]
\includegraphics[width=0.8\linewidth]{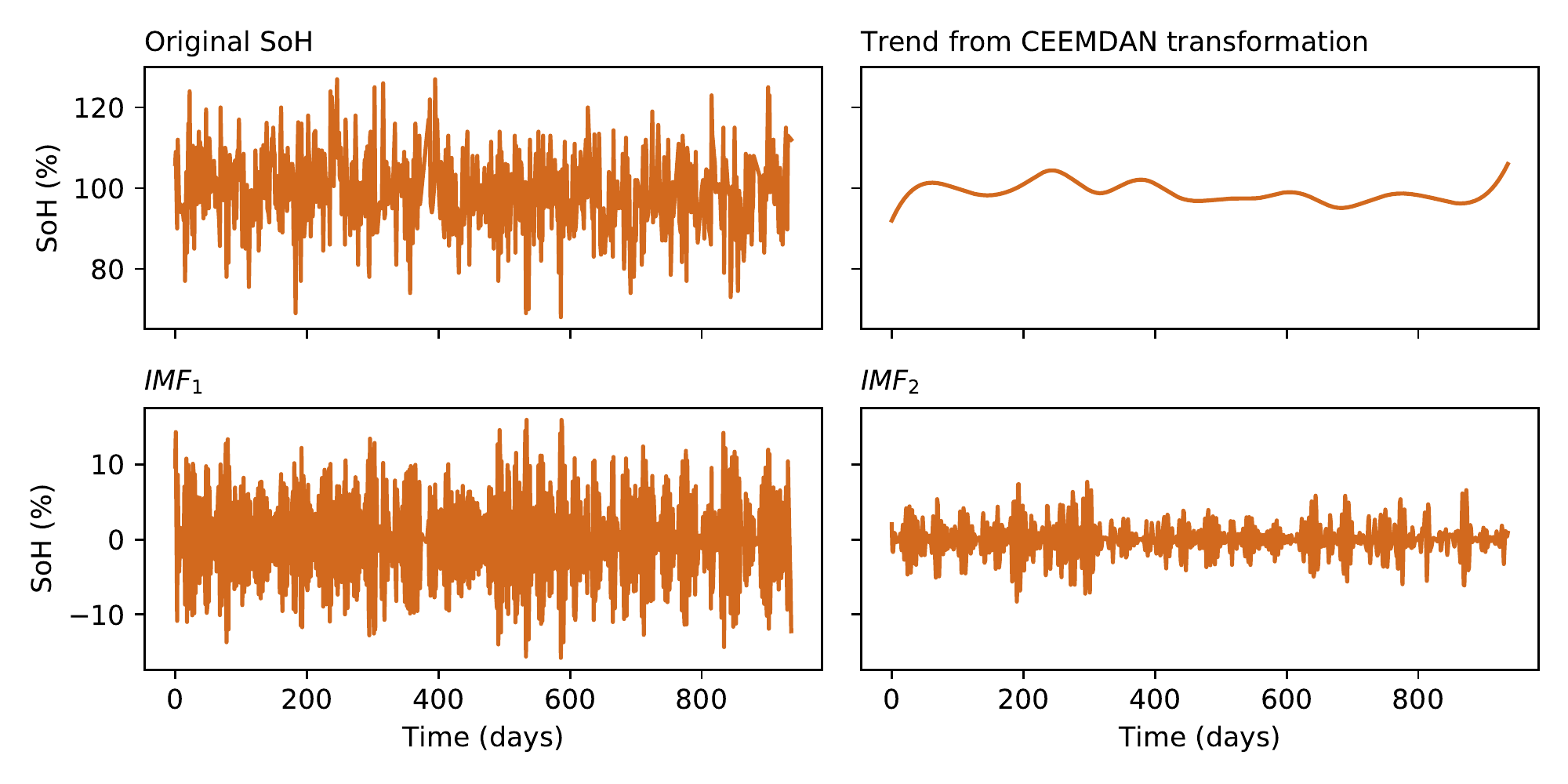}
\caption{SoH, trend and two first IMFs obtained from complete empirical mode decomposition transformation for battery (a).}
\label{fig:imfs_a}
\end{figure*}

From this initial data the outliers were detected and removed by the interquartile range selection method; any observations that were more than 1.5 $\times$ IQR below the Q1 or more than 1.5 $\times$ IQR above the Q3 were considered as outliers \cite{zwillinger1999crc}; some 3--5\% of the data was discarded. Then the missing daily values were imputed by taking the average of the values before and after the missing values. As the next step for the model generation, the step 3 in Figure \ref{fig:data_model}, we selected the complete empirical mode decomposition (CEEMDAN) implementation \cite{luukko2016introducing}, which fundamentally implements the method as described in the methodology section \ref{methodology} (an example of decomposition's results is in Figure \ref{fig:imfs_a}). To obtain the instantaneous frequency for the SoH, we performed the Huang-Hilbert transform at this stage (example  in Figure \ref{fig:ifreq}), lagged it by one time step (t-1), and added it as a predictor. The summary of the features after the feature extraction is in Table \ref{table:predictors}. 

\begin{table*}[hbt!]
\centering
\scriptsize
\caption{Selected features used by a model. The ticks represent the features used by a model.}
\begin{tabular}{|p{5.6cm}|c|c|c|}
\hline
\textbf{Signals/ Target (predictors)} & \textbf{IMFs} & \textbf{SoH (basic)} & \textbf{SoH (basic+IMFs)}                                      \\ \hline
Time stamp of the data   & \cmark & \cmark & \cmark  \\ \hline
Voltage V  & \cmark & \cmark & \cmark                                       \\ \hline
Current A  & \cmark & \cmark & \cmark                                       \\ \hline
SOC \%  & \cmark & \cmark & \cmark                                                \\ \hline
Ambient temperature $^{\circ}$C & \cmark & \cmark & \cmark    \\ \hline

Charging length in minutes  & \cmark & \cmark & \cmark \\ \hline
Energy Wh  & \cmark & \cmark & \cmark                                  \\ \hline
Voltage difference $\Delta$V & \cmark & \cmark & \cmark   \\ \hline

Cycle    & \cmark& \cmark& \cmark \\ \hline
(t-1) lagged instantaneous frequency & \xmark & \cmark & \cmark   \\ \hline
Prediction for residue  & N/A & \xmark & \cmark               \\ \hline
Prediction for  IMFs & N/A & \xmark & \cmark                \\ \hline
\end{tabular}
\label{table:predictors}
\end{table*}

For the IMF and residue predictions, in the model tuning and validation phases (steps 4 -- 5 in Figure \ref{fig:data_model}), 70\% of the data set was used  for tuning and 30\% for validation. We selected XGB randomly as the method to use, albeit proven one in the field of batteries \cite{koz03ele, SEPASI2014368}. For the tuning, we used the predictors presented in Table \ref{table:predictors} and tuned the model using 10-fold timeseries cross-validation; the tuning results are presented in Table \ref{table:emdhybridparams}. The best models were verified with the test set. Finally, the resulted predictions for each of the IMFs and for the residue were added to the final data set as additional predictors maintaining the train and test split for avoiding data leakage.

\subsection{Model tuning and initial model comparison for predicting the SoH of a battery pack} \label{model-tuning-1comparison}

Next, we generated the models for predicting SoH (steps 6--7 in Figure \ref{fig:data_model}). We selected seven different up-to-date methods to ensure that we find a good model for the prediction. We tuned the GB, LGB, RF, ETR and XGB models with the predictors presented in the Table \ref{table:predictors}. For this purpose,  we re-framed the timeseries to windows with size seven (six past observations and prediction horizon of one) and utilized 10-fold timeseries cross-validation in the grid search of the best hyperparameters. For each of the methods, we tuned two models with SoH as the target: the first model utilised basic predictors and the second utilised basic predictors together with intrinsic mode function (IMF) and residue prediction values (i.e., S\~{o}H$(t)$). The summary of the tuned hyperparameters is provided in Table \ref{table:sohhybridparams}.

Furthermore, two simple LSTM and CNN-LSTM models were grid searched (network structure in Table \ref{table:nnhybridparams}) for finding the best hyperparameters for them. In the evaluation phase, we normalised the used data (each predictor variable and the target variable), and repeated each model evaluation 10 times and averaged the results after inverting the normalisation for the final results. All of the predictors were used for these models (Table \ref{table:predictors}). This grid search method was used as an initial attempt to yield a network structure for this data set as no prefitted neural network model exists for this data.

\begin{table}[hbt!]
\centering
\scriptsize
\caption{Overview of the selected models for predicting intrinsic mode function and residue values.}
\begin{tabular}{|l|l|l|l|l|l|}
\hline
\multicolumn{2}{|l|}{\textbf{Target}}      & \textbf{IMF1} & \textbf{IMF2} & \textbf{IMF3} & \textbf{residue} \\ 
\hline
\multicolumn{2}{|l|}{Method}               & XGB           & XGB           & XGB           & XGB              \\ \hline
\multicolumn{2}{|l|}{Number of estimators} & 1000          & 500           & 250           & 250              \\ \hline
\multicolumn{2}{|l|}{Maximum depth}        & 2             & 2             & 2             & 3                \\ \hline
\multirow{3}{*}{Column samples} & by level & 0.8           & 0.8           & 0.8           & 0.8              \\ \cline{2-6} 
                                & by node  & 0.8           & 0.5           & 0.5           & 0.8              \\ \cline{2-6} 
                                & by tree  & 0.8           & 0.8           & 0.8           & 0.5
\\ \hline
\end{tabular}
\label{table:emdhybridparams}
\end{table}

\begin{table*}[hbt!]
\centering
\scriptsize
\caption{Overview of the selected regression models for predicting SoH (window size 7).}
\begin{adjustbox}{max width=\textwidth}
\begin{tabular}{|l|l|l|l|l|l|l|l|l|l|l|l|}
\hline
\multicolumn{2}{|l|}{\textbf{Target}}      & \textbf{SoH} & \textbf{SoH} & \textbf{SoH} & \textbf{SoH} & \textbf{SoH} & \textbf{SoH} & \textbf{SoH} & \textbf{SoH} & \textbf{SoH} & \textbf{SoH} \\ 
\hline
\multicolumn{2}{|l|}{Method}               & GB           & EMD-GB       & LBG          & EMD-LGB      & RF           & EMD-RF       & XGB          & EMD-XBG      & ETR          & EMD-ETR      \\ \hline
\multicolumn{2}{|l|}{Number of estimators} & 1000         & 1000         & 250          & 500          & 500          & 1000         & 1000         & 1000         & 500          & 1000         \\ \hline
Maximum depth                   &          & 4            & 4            & 8            & 6            & 6            & 5            & 2            & 2            & 6            & 6            \\ \hline
\multirow{3}{*}{Column samples} & by level & N/A          & N/A          & N/A          & N/A          & N/A          & N/A          & 0.8          & 0.8          & N/A          & N/A          \\ \cline{2-12} 
                                & by node  & N/A          & N/A          & N/A          & N/A          & N/A          & N/A          & 0.8          & 0.8          & N/A          & N/A          \\ \cline{2-12} 
                                & by tree  & N/A          & N/A          & 0.8          & 0.8          & N/A          & N/A          & 0.8          & 0.8          & N/A          & N/A          \\ \hline
\multicolumn{2}{|l|}{Subsamples}           & 0.8          & 0.8          & 0.8          & 0.8          & N/A          & N/A          & 1            & 1            & N/A          & N/A          \\ \hline
\multicolumn{2}{|l|}{Subsample frequency}  & N/A          & N/A          & 5            & 5            & N/A          & N/A          & N/A          & N/A          & N/A          & N/A          \\ \hline
\end{tabular}
\end{adjustbox}
\label{table:sohhybridparams}
\end{table*}

\begin{table*}[hbt!]
\centering
\scriptsize
\caption{Overview of the LSTM and CNN-LSTM models for predicting SoH (window size 7).}
\begin{adjustbox}{max width=\textwidth}

\begin{tabular}{|lll|l|l|l|}
\hline
\multicolumn{3}{|l|}{\textbf{LSTM}} & \multicolumn{3}{l|}{\textbf{CNN-LSTM}}           \\ \hline
\multicolumn{1}{|l|}{\textbf{\begin{tabular}[c]{@{}l@{}}Layer\\ (type)\end{tabular}}} & \multicolumn{1}{l|}{\textbf{Output shape}} & \textbf{\begin{tabular}[c]{@{}l@{}}Number \\ of parameters\end{tabular}} & \textbf{\begin{tabular}[c]{@{}l@{}}Layer\\ (type)\end{tabular}} & \textbf{Output shape} & \textbf{\begin{tabular}[c]{@{}l@{}}Number \\ of parameters\end{tabular}} 
\\ \hline
\multicolumn{1}{|l|}{lstm\_10 (LSTM)}                                                 & \multicolumn{1}{l|}{(None, 200)}           & 172800                                                                   & conv1d\_2 (Conv1D)                                              & (None, 12, 64)        & 2944                                                                     \\ \hline
\multicolumn{1}{|l|}{repeat\_vector\_5 (RepeatVecto}                                  & \multicolumn{1}{l|}{(None, 7, 200)}        & 0                                                                        & conv1d\_3 (Conv1D)                                              & (None, 10, 64)        & 12352                                                                    \\ \hline
\multicolumn{1}{|l|}{lstm\_1 (LSTM)}                                                  & \multicolumn{1}{l|}{(None, 7, 200)}        & 416800                                                                   & max\_pooling1d\_1 (MaxPooling1                                  & (None, 5, 64)         & 0                                                                        \\ \hline
\multicolumn{1}{|l|}{time\_distributed\_2 (TimeDist}                                  & \multicolumn{1}{l|}{(None, 7, 50)}         & 10050                                                                    & flatten\_1 (Flatten)                                            & (None, 320)           & 0                                                                        \\ \hline
\multicolumn{1}{|l|}{time\_distributed\_3 (TimeDist}                                  & \multicolumn{1}{l|}{(None,7,1)}            & 51                                                                       & repeat\_vector\_1 (RepeatVecto                                  & (None, 7, 320)        & 0                                                                        \\ \hline
                                                                                      &                                            &                                                                          & lstm\_1 (LSTM)                                                  & (None, 7, 200)        & 416800                                                                   \\ \cline{4-6} 
                                                                                      &                                            &                                                                          & time\_distributed\_2 (TimeDist                                  & (None, 7, 50)         & 10050                                                                    \\ \cline{4-6} 
                                                                                      &                                            &                                                                          & time\_distributed\_3 (TimeDist                                  & (None,7,1)            & 51                                                                       \\ \cline{4-6} 
\hline
\end{tabular}
\end{adjustbox}
\label{table:nnhybridparams}
\end{table*}

For the five methods with the best EVAR yielded (in Table \ref{table:modelcomp}), we expanded the number of models. We utilised the same methods but increased the number of window sizes utilised. For each new window size, we tuned a model of its own.  After tuning, we made a comparison between all the yielded models with the novel algorithm \ref{alg:wf-pci} for finding the optimal values for samples sizes, rolling window sizes and roll step sizes for the models for this data set (subsection \ref{wf-battery-evaluation}). 

\subsection{Verification of the novel walk-forward algorithm and calculating the point-wise confidence intervals with an external data set} \label{wf-evaluation}

\begin{figure}[hbt!]
\centering
\includegraphics[width=0.5\linewidth]{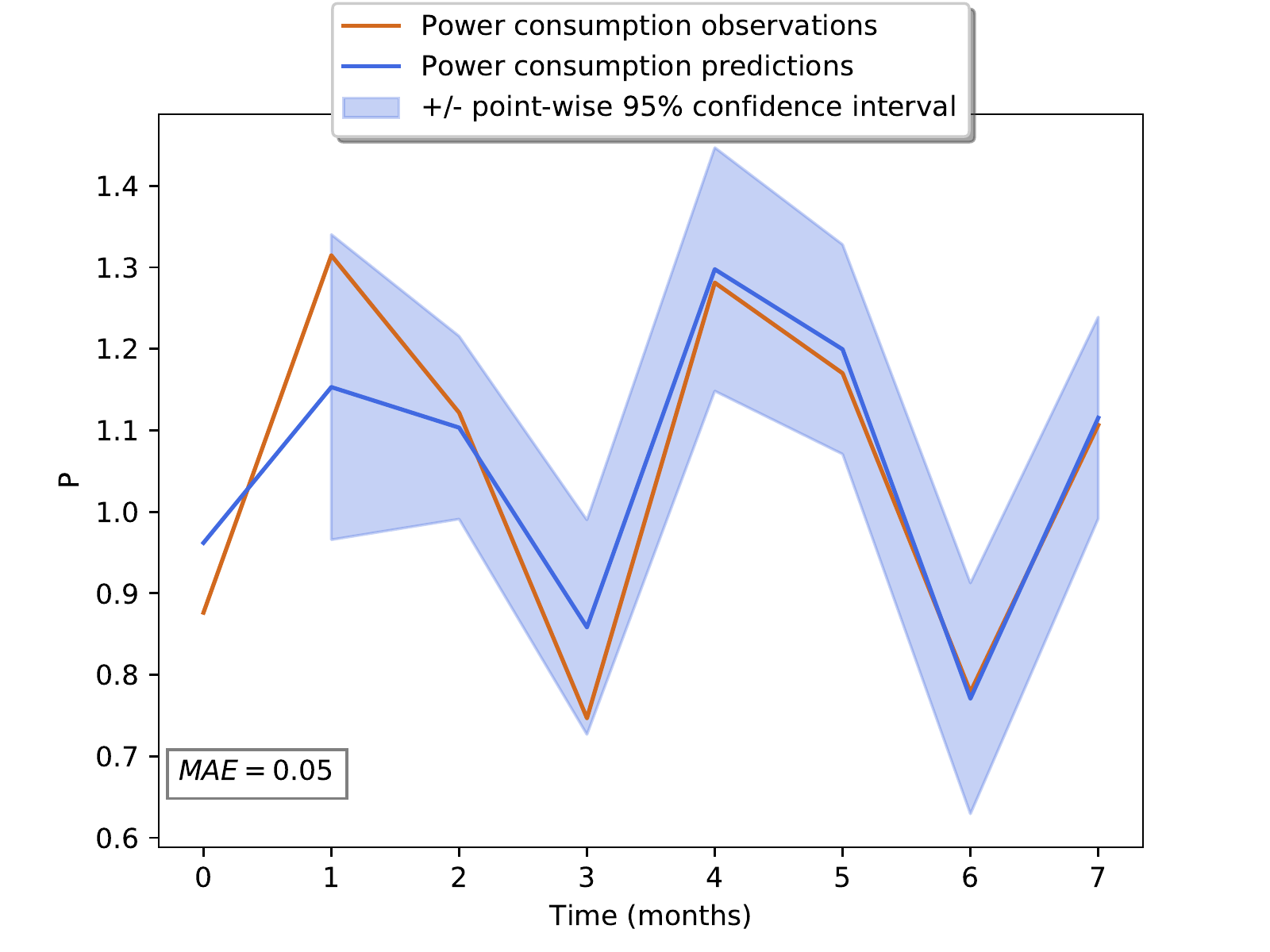}
\caption{The graph of the household power consumption predictions with the XGB model and the point-wise CI for the predictions. Notice that the confidence interval varies due to, e.g., the used aggregation method, the rolling window size and the underlying model's stability.}
\label{fig:house_all}
\end{figure}

 We verified the novel walk-forward method for yielding the point-wise confidence intervals against a public data set (the household power consumption data set \cite{Hebrail2010}); this data was aggregated to monthly values. The model that was used to predict the monthly power consumption was a simple XGB model (number of estimators: 50, maximum depth: 2); it yielded the RMSE 0.08, which exceed the na\"ive model's RMSE 0.11. For the verification, the number of observation windows used for walk-forward was 30, the size of the sliding window was 7 and the number of rolled over windows was 4. The yielded prediction results with the corresponding point-wise confidence intervals are in Figure \ref{fig:house_all}. The CI does not fluctuate significantly nor show a clear trend; the model seems to be stabile \cite{MATLAB2021}. As these results supported the theoretical basis \cite{has17elewf} for its use as a verification method, the point-wise confidence interval calculation method was used in our model development as well.

\subsection{Verification of the novel walk-forward method and calculating the point-wise confidence intervals with battery data set} \label{wf-battery-evaluation}

For the finally selected methods (in section  \ref{model-tuning-1comparison}), we used the predictors presented in Table \ref{table:predictors} and tuned the models using 10-fold timeseries cross-validation; the tuning results are presented in Table \ref{table:w14_91hybridparameters}. 
After the tuning, the yielded models were verified with the novel algorithm with both expanding and sliding windows. As a summary, the following sizes were utilised for samples: 14, 30 and 90 days, for windows: 7, 14 and 30 days (with prediction horizon 1 included in these figures), and for window rolls: 1, 2, 7, and 14 days. In order to make enough repetitions, the algorithm parameters were set so that sample size $>$ window size $>$ 2 x roll size, except for one case where roll and window sizes were set to be equal. It can be noted that sample size determines the initial test set, from which walk-forward starts to roll the window towards the newest observations in the test set. 

\begin{table*}[hbt!]
\centering
\scriptsize
\caption{Overview of the further selected regression models for predicting SoH (window sizes 14--91).}
\begin{adjustbox}{max width=\textwidth}
\begin{tabular}{|c|l|l|l|l|l|l|l|l|}
\hline
\multicolumn{4}{|l|}{\textbf{Target}}                                                                           & \textbf{SoH} & \textbf{SoH} & \textbf{SoH} & \textbf{SoH} & \textbf{SoH} \\ \hline
\multicolumn{4}{|l|}{Method}                                                                                    & GB           & EMD-GB       & RF           & EMD-RF       & EMD-ETR      \\ \hline
\multicolumn{1}{|l|}{\multirow{3}{*}{Window}} & \multirow{3}{*}{14} & \multicolumn{2}{l|}{Number of estimators} & 250          & 250          & 250          & 1000         & 1000         \\ \cline{3-9} 
\multicolumn{1}{|l|}{}                        &                     & \multicolumn{2}{l|}{Maximum depth}        & 4            & 4            & 5            & 6            & 6            \\ \cline{3-9} 
\multicolumn{1}{|l|}{}                        &                     & \multicolumn{2}{l|}{Subsamples}           & 0.8          & 0.8          & N/A          & N/A          & N/A          \\ \hline
\multirow{3}{*}{Window}                       & \multirow{3}{*}{30} & \multicolumn{2}{l|}{Number of estimators} & 1000         & 1000         & 500          & 500          & 500          \\ \cline{3-9} 
                                              &                     & \multicolumn{2}{l|}{Maximum depth}        & 5            & 5            & 5            & 6            & 6            \\ \cline{3-9} 
                                              &                     & \multicolumn{2}{l|}{Subsamples}           & 0.8          & 0.8          & N/A          & N/A          & N/A          \\ \hline
\multirow{3}{*}{Window}                       & \multirow{3}{*}{91} & \multicolumn{2}{l|}{Number of estimators} & 1000         & 1000         & 250          & 250          & 500          \\ \cline{3-9} 
                                              &                     & \multicolumn{2}{l|}{Maximum depth}        & 4            & 4            & 5            & 5            & 6            \\ \cline{3-9} 
                                              &                     & \multicolumn{2}{l|}{Subsamples}           & 0.8          & 0.8          & N/A          & N/A          & N/A          \\ \hline
\end{tabular}
\end{adjustbox}
\label{table:w14_91hybridparameters}
\end{table*}

\section{Results and discussion} \label{results}

\subsection{Results for battery pack (a) and comparison with the previous methods} \label{wf-comparison}

\begin{table*}[hbt!]
\centering
\scriptsize
\caption{Summary of the model comparisons (EVAR is the explained variance). The five best grid searched models are in boldface.}
\begin{tabular}{|l|l|l|l|l|l|l|}
\hline
\textbf{Battery} & \textbf{model} & \textbf{samples} & \textbf{EVAR} & \textbf{MAE} & \textbf{RMSE} \\ \hline
\textbf{Battery (a)}      & \textbf{GB}             & \textbf{70-99\% of data } & \textbf{99.9}          & \textbf{0.21}         & \textbf{0.28}          \\ \hline
\textbf{Battery (a)}      & \textbf{EMD-GB}         & \textbf{70-99\% of data}  & \textbf{99.9}          & \textbf{0.20}         & \textbf{0.28}        \\ \hline
\textbf{Battery (a)}      & \textbf{RF}             & \textbf{70-99\% of data}  & \textbf{99.9}          & \textbf{0.22}         & \textbf{0.31}          \\ \hline
\textbf{Battery (a)}      & \textbf{EMD-RF}       & \textbf{70-99\% of data}  & \textbf{99.9}          & \textbf{0.24}        & \textbf{0.32}         \\ \hline
Battery (a)      & LGB           & 70-99\% of data  & 99.2          & 0.48         & 0.81          \\ \hline
Battery (a)      & EMD-LGB       & 70-99\% of data  & 99.3          & 0.45         & 0.75          \\ \hline
Battery (a)      & EXT            & 70-99\% of data  & 99.6          & 0.51         & 0.59          \\ \hline
\textbf{Battery (a)}      & \textbf{EMD-EXT}        & \textbf{70-99\% of data}  & \textbf{99.7}        & \textbf{0.43}         & \textbf{0.53}  \\ \hline
Battery (a)      & XGB            & 70-99\% of data  & 99.3          & 0.53         & 0.76          \\ \hline
Battery (a)      & EMD-XGB      & 70-99\% of data  & 99.6          & 0.44         & 0.59          \\ \hline
Battery (a)      & EMD-LSTM       & 70\% of data     & -             & 7.70         & 9.47          \\ \hline
Battery (a)      & EMD-CNN LSTM   & 70\% of data     & -             & 8.09         & 10.27         \\ \hline

\end{tabular}
\label{table:modelcomp}
\end{table*}

The best initial model for the battery (a) was yielded by gradient boosting with additional predictors (i.e., S\~{o}H$(t)$) and  with window size 7. (See (EMD-GB) and the rest of the results in Table \ref{table:modelcomp}). This model yielded mean absolute error 0.20 and RMSE of 0.28 that are in the low error ranges. Furthermore, it can be noted that all of the four best regression models were close-by each other in terms of MAE. For the LSTM and LSTM-CNN it can be noted that the results indicate that an optimal neural network structure and parameters were not found this time. 

In the further validation, we tuned new models, and then we back-tested them with the novel walk-forward method using the battery (a). The results (two of the best scores and the worst score for each model) are presented in Table \ref{table:rolled_wf_batteries}.
The best initial model for the battery (a) was gradient boosting without the additional predictors (i.e., without S\~{o}H$(t)$) and with window size 14. 

\begin{table*}[hbt!]
\centering
\scriptsize
\caption{The two best results and the worst result for SoH predictions for the models with S\~{o}H$(t)$ predictors (EMD-) and without them for battery (a). All models are verified with both expanding and sliding  window. The best model yielded, is in boldface.}
\begin{tabular}{|l|l|l|l|l|l|l|l|l|l|l|}
\hline
\textit{\textbf{roll type}} & \multicolumn{5}{l|}{\textit{expanding window}}                                 & \multicolumn{5}{l|}{\textit{sliding window}}                                  \\ \hline
\textbf{model}              & \textbf{sample} & \textbf{win} & \textbf{roll} & \textbf{MAE}  & \textbf{RMSE} & \textbf{sample} & \textbf{win} & \textbf{roll} & \textbf{MAE} & \textbf{RMSE} \\ \hline
\multirow{3}{*}{GB}         & \textbf{30}     & \textbf{14}  & \textbf{1}    & \textbf{0.18} & \textbf{0.23} & 14              & 7            & 1             & 0.20         & 0.24          \\ \cline{2-11} 
                            & 30              & 14           & 2             & 0.19          & 0.23          & 30              & 7            & 1             & 0.20         & 0.25          \\ \cline{2-11} 
                            & 14              & 7            & 2             & 0.26          & 0.33          & 14              & 7            & 2             & 0.25         & 0.33          \\ \hline
\multirow{3}{*}{EMD-GB}     & 30              & 14           & 2             & 0.19          & 0.23          & 30              & 14           & 1             & 0.20         & 0.25          \\ \cline{2-11} 
                            & 90              & 30           & 1             & 0.20          & 0.22          & 90              & 14           & 1             & 0.21         & 0.25          \\ \cline{2-11} 
                            & 14              & 7            & 2             & 0.26          & 0.33          & 90              & 30           & 7             & 0.25         & 0.32          \\ \hline
\multirow{3}{*}{RF}         & 30              & 7            & 2             & 0.20          & 0.27          & 30              & 7            & 2             & 0.20         & 0.27          \\ \cline{2-11} 
                            & 30              & 7            & 1             & 0.22          & 0.27          & 30              & 7            & 1             & 0.22         & 0.28          \\ \cline{2-11} 
                            & 90              & 30           & 7             & 0.29          & 0.39          & 90              & 30           & 2             & 0.29         & 0.45          \\ \hline
\multirow{3}{*}{EMD-RF}     & 30              & 14           & 1             & 0.21          & 0.27          & 30              & 14           & 2             & 0.20         & 0.27          \\ \cline{2-11} 
                            & 30              & 14           & 2             & 0.21          & 0.27          & 30              & 14           & 1             & 0.21         & 0.27          \\ \cline{2-11} 
                            & 90              & 30           & 7             & 0.30          & 0.42          & 14              & 7            & 2             & 0.29         & 0.33          \\ \hline
\multirow{3}{*}{EMD-ETR}    & 90              & 30           & 7             & 0.39          & 0.49          & 90              & 30           & 7             & 0.41         & 0.50          \\ \cline{2-11} 
                            & 90              & 7            & 1             & 0.46          & 0.55          & 30              & 14           & 1             & 0.42         & 0.51          \\ \cline{2-11} 
                            & 14              & 7            & 2             & 0.54          & 0.63          & 14              & 7            & 2             & 0.57         & 0.66          \\ \hline
\end{tabular}
\label{table:rolled_wf_batteries}
\end{table*}

This model yielded mean absolute error 0.18 and RMSE of 0.20, and is the final best result for this paper. Furthermore, for gradient boosting and for random forests models, the overall difference between the best and the worst MAE for all of the submodels was narrow ($<$ 0.1). The best models were yielded without the additional predictors (i.e., without S\~{o}H$(t)$). In contrast to these results, for the light gradient boosting, extreme boosting and extra trees, the overall difference between the best and the worst MAE for all of the submodels was typically wider than 0.1. The best models were yielded utilising the additional predictors (i.e., with S\~{o}H$(t)$). A conclusion is that the decomposed S\~{o}H$(t)$ predictions used as predictors slightly enhanced some models' overall prediction accuracy. The expanding  window  method yields, in majority of the cases,  slightly smaller MAE and RMSE values for this data set than the sliding window method. This is according to the general findings in the industry for  data with relatively few  samples \cite{uber2021slideexpand}. The window roll sizes 1-7 yielded models that had the best and the worse MAE results close by each other with the same  window size (e.g. 0-0.08 difference in MAE in Table \ref{table:rolled_wf_batteries}), i.e., a roll size of 7 can be applied to this data set without affecting the MAE results. Moreover, we spot-tested some roll sizes that do not have overlap with the previous window (e.g. window of size 14 and window roll of size 14); however, the MAE deviated in random manner, and in many cases by 50\% from the results with the same window size but with a smaller roll. This indicates that a window roll that is up to 23-28\% of the utilised window size is applicable to this data set for yielding reliable results.

The best model (GB) was refitted to battery (a), after which we made predictions anew (Figure \ref{fig:ci120}). We yielded the MAE loss function value of 1.52 and the goodness of the fit, $R^2$, 0.91; the model outperformed the ARIMA model that we introduced in our previous paper (Table \ref{table:comparative}), although neither of these models perform well over the entire 32-month period.

Furthermore, as this timeseries model development setup was designed to predict SoH for the near future (sample sizes used were the newest 14-90 days and window sizes were 7-30 days), we refitted the model to the 3 nearest months (Figure \ref{fig:ci12}). We yielded the MAE loss function value of 0.21  and the goodness of the fit, $R^2$, 1, which indicates that the model predicts well over three-month-period. 
Moreover, it should be noted that the model can be overconfident in its predictions indicated by a $R^2$ score that is one. Therefore, in order to set confidence intervals for estimating model stability, the novel walk-forward method was applied to yield point-wise confidence intervals. (See subsection \ref{pointwise_ci}).

\begin{table}[hbt!]
\centering
\scriptsize
\caption{The GB model comparative analysis with previous ARIMA model results \cite{huo20dyn}}
\begin{tabular}{|l||l|l|l|}
\hline
\textbf{} & \textbf{GB model} & \textbf{ARIMA} \\ \hline

RMSE      & 1.56          & 2.68         \\ \hline
R2        & 0.91          & -0.26          \\ \hline
\end{tabular}
\label{table:comparative}
\end{table}

\subsection{Results for a set of batteries} \label{results batteryset}

For testing the timeseries of all batteries, we evaluated them with the Wilcoxon signed-rank test \cite{wilcoxon, mak18sta}. The test results were used to assess if a uniform forecast model can be applied to the set of batteries or not. Our hypothesis was that we can apply our model to the set of battery packs, as they come from the same factory, have the same calendar age and are used in the similar forklifts. The hypothesis (H0) was set as follows: the sample distributions from different batteries were related to the battery (a). Wilcoxon yielded that 65\% of the batteries had the same distribution as battery (a) (failed to reject H0), and consequently 35\% had different distribution (rejected H0).

\begin{table}[hbt!]
\centering
\scriptsize
\caption{Evaluation results of verifying the GB model with data from batteries b-e.}
\begin{tabular}{|l|l|l|l|l|l|}
\hline
\textbf{Battery}     & \textbf{(b)} & \textbf{(c)} & \textbf{(d)} & \textbf{(e)} & \textbf{(f)} 
\\ \hline
EVAR                 & 99.9\%       & 99.8\%       & 98.8\%       & 99.8\%   &  99.9\%   \\ \hline
MAE                  & 0.18         & 0.29         & 0.20         & 0.15 & 0.27      \\ \hline
RMSE                 & 0.26         & 0.39         & 0.90         & 0.27 & 0.30      \\ \hline
R$^{2}$ & 1.0         & 1.0         & 0.99  & 1.0  & 1.0         \\ \hline
\end{tabular}
\label{table:evaluate_bcdef}
\end{table}

\begin{figure*}[hbt!]
\centering
\includegraphics[width=0.8\linewidth]{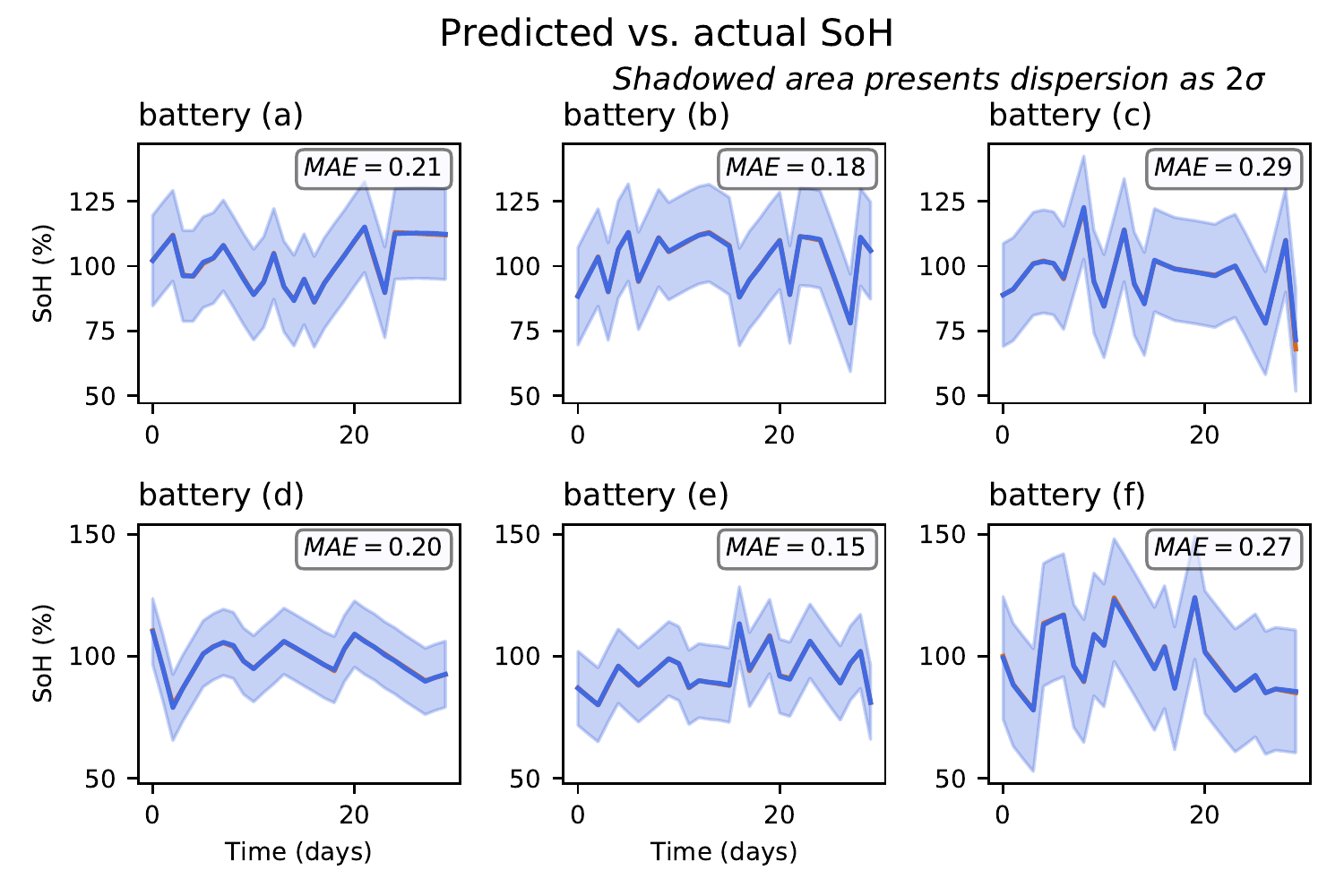}
\caption{Battery (b\textendash f) GB predicted SoH values. The blue line is the predicted and the orange is the observed SoH. The shadowed area is +/- two standard deviations.}
\label{fig:abcde}
\end{figure*}

Amongst the batteries in the same distribution,
the batteries (b\textendash f) were scrutinised in more detailed manner. For the verification, the best model extracted for battery (a) was applied to the five battery packs. The verification yielded MAE between 0.15\textendash 0.29 and goodness of the fit, R$^2$, between 0.99\textendash 1.00 (Table \ref{table:evaluate_bcdef} and Figure \ref{fig:abcde}).  The overall evaluation results for the set of batteries are promising; however, there is a need for further analysis on, e.g., the environmental factors and the length of the battery data on the battery SoH forecast; these may make a difference for enhancing the model and its  reliability. 

\subsection{Results for calculating the point-wise confidence intervals for battery (a)}\label{pointwise_ci}

\begin{figure*}[hbt!]
\centering
\includegraphics[width=0.6\linewidth]{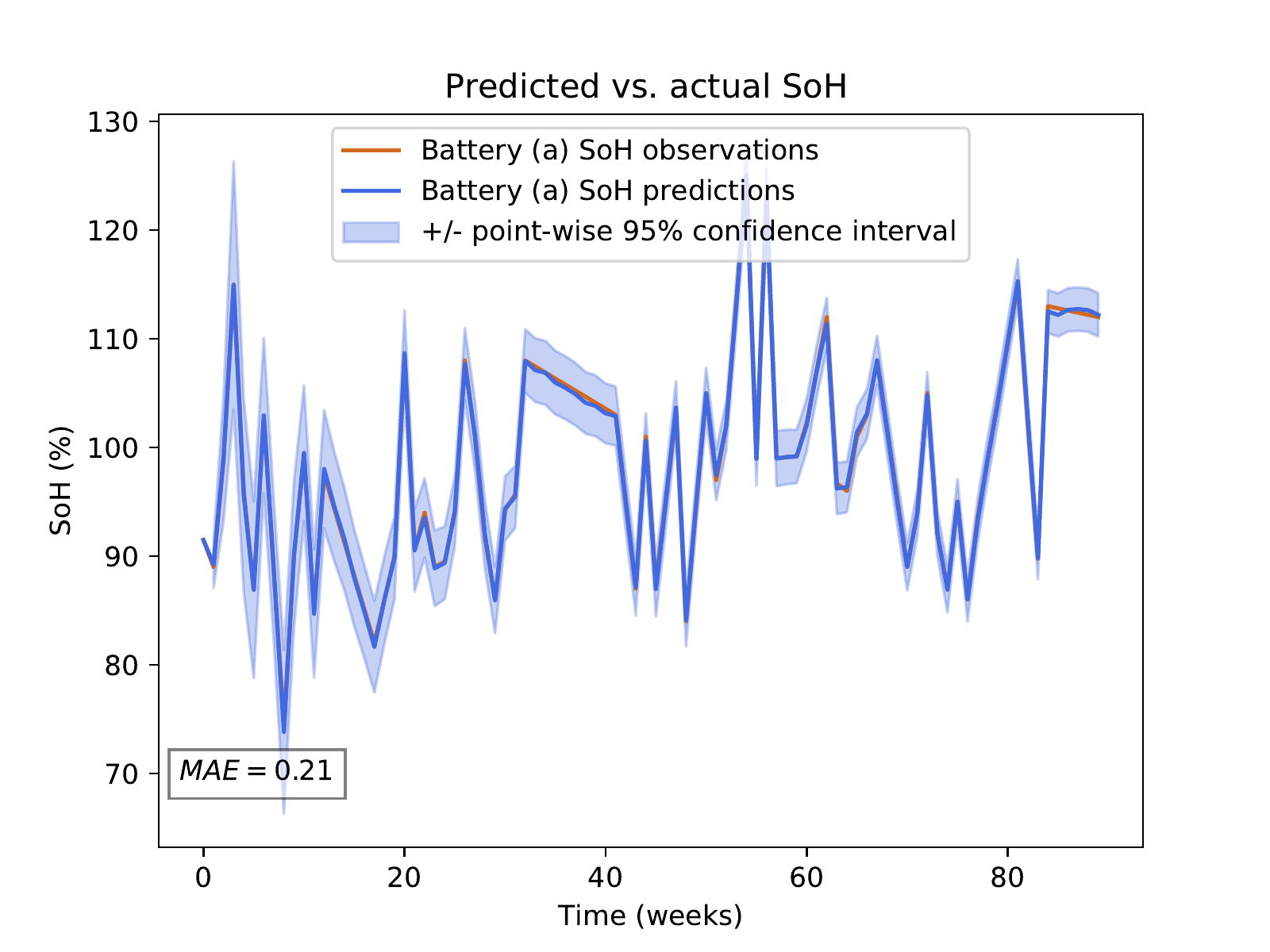}
\caption{Battery (a) EMD-LGB model predicted SoH values for three months (90 days) with confidence interval. The blue line is the predicted and the orange is the observed SoH. The shadowed area indicated point-wise 95\% confidence interval for predictions.}
\label{fig:ci12}
\end{figure*}

\begin{figure*}[hbt!]
\centering
\includegraphics[width=0.6\linewidth]{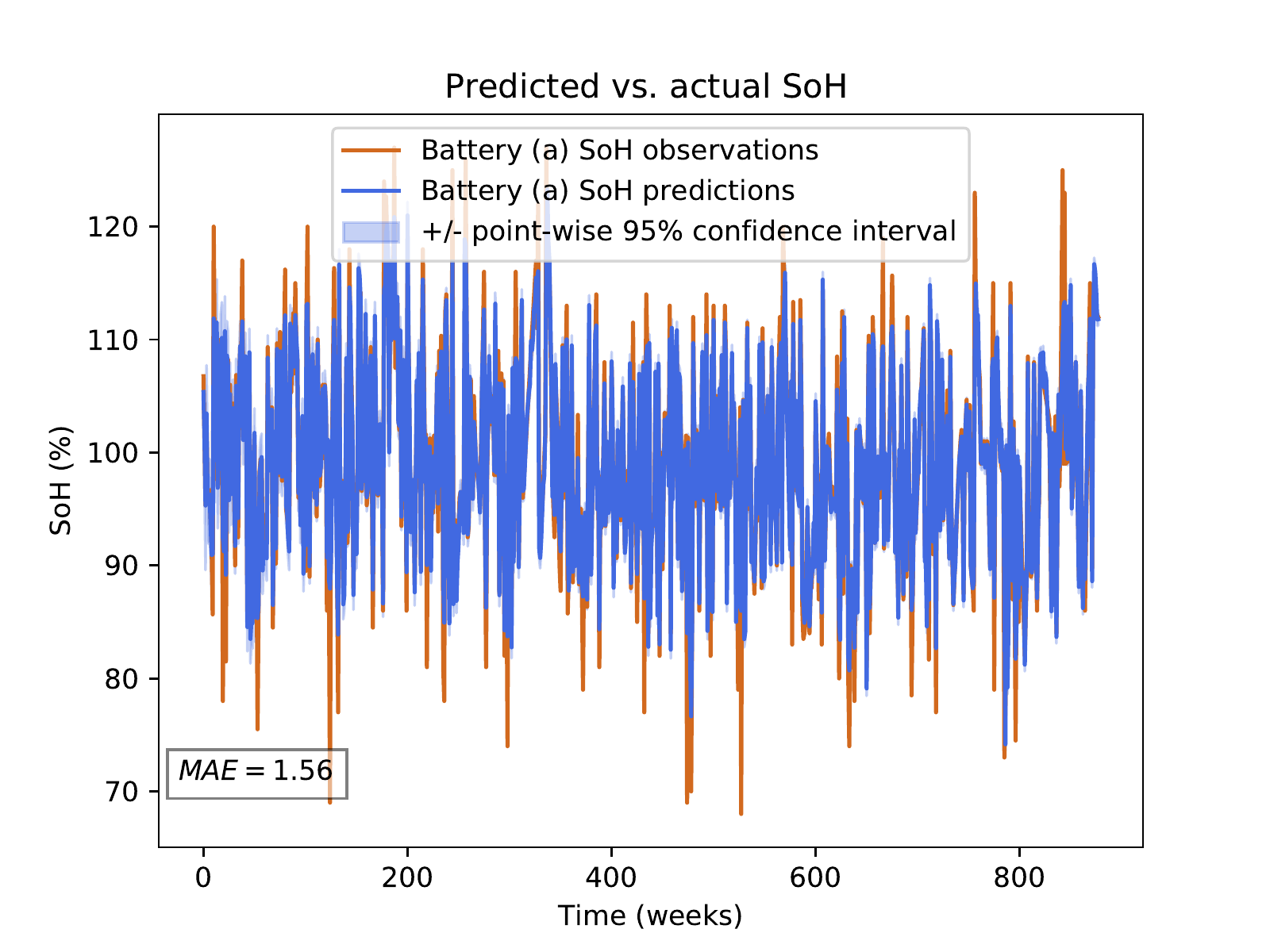}
\caption{Battery (a) SoH values for 30 months with the confidence interval; the blue line is the predicted and the orange is the observed SoH. The shadowed area indicates point-wise 95\% confidence interval for predictions.}
\label{fig:ci120}
\end{figure*}

We calculated the confidence intervals for the SoH predictions for battery (a) in order to evaluate the overall model behavior. As depicted in Figures \ref{fig:ci12} and \ref{fig:ci120}, we used sample size of 3 and 30 months; we refitted the best GB model to the data and depicted the results. For the 3-month sample size the point-wise confidence intervals varied so that, at its narrowest, the range 88.0--91.9 covered the true prediction with the 95\% likelihood and, at it widest, the range 103.6--126.6  covered the true prediction with the 95\% likelihood; overall, for a 3-month period this GB model yielded reasonable predictions.

As a final remark on the results, calculating the predictions with confidence intervals with data re-split and model refit at every walk-forward iteration step is computationally heavy; however for the relatively small number of observations (around 1000 for 3-year period and 3700 for the 10-year period for one battery pack), this is tolerable; furthermore the window roll diminishes the number of calculations.

\subsection{Results for instantaneous frequency}\label{pointwise_cix}

\begin{figure*}[hbt!]
\centering 
\includegraphics[width=0.6\linewidth]{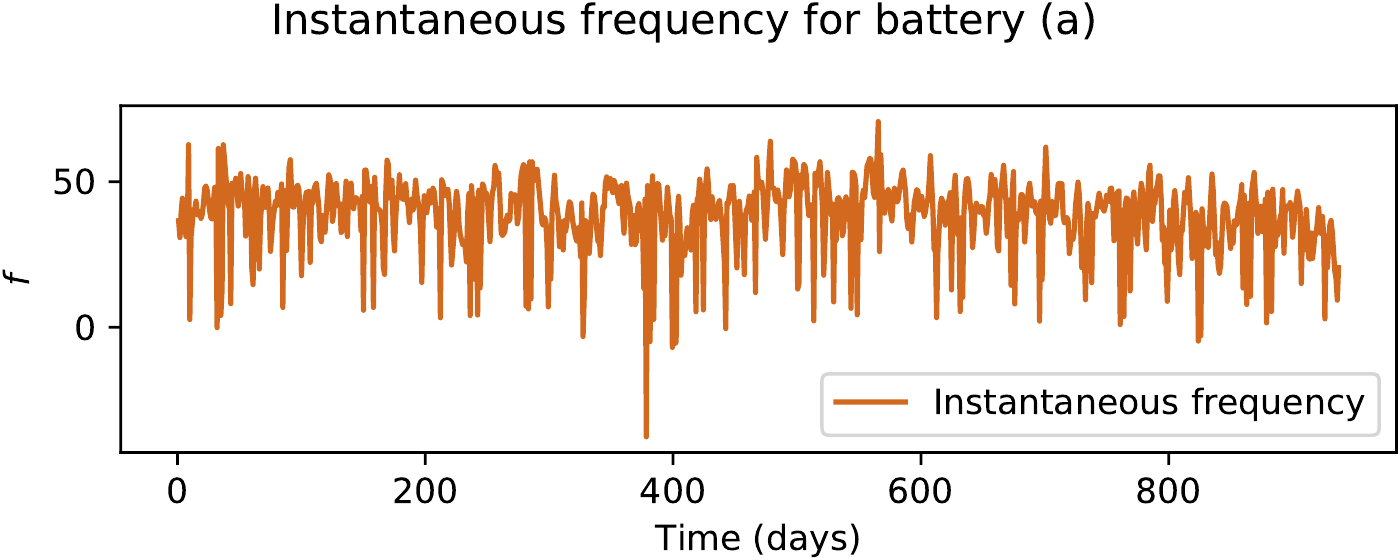}
\caption{The used Hilbert-Huang transformed instantaneous frequency for battery (a).}
\label{fig:ifreq}
\end{figure*}

As a second but last result for this paper, the analytic signal derived from the decomposed SoH provides means to detect changes i.e., it provides means to analyse if the SoH starts to deteriorate, or, to change less frequently or more frequently than before; this information alleviates the decision to initiate an inspection of the battery in the field, or, to change the model. For example, the persistent drop in the instantaneous frequency (Equation \ref{eq:ifreq}) around the day 900 may indicate a general trend that may require revamping the model if this change is permanent (Figure \ref{fig:ifreq}). Furthermore, there is a peak around day 380 in the figure. Inspecting the corresponding data file revealed that the SoH values fluctuated 30\% between consequent days during several days; this indicated some kind of transient failure, and revealing its root cause would require a further inspection. 

\subsection{Results for the equivalent cycles} \label{feature results}

\begin{figure*}[hbt!]
\centering
\includegraphics[width=0.6\linewidth]{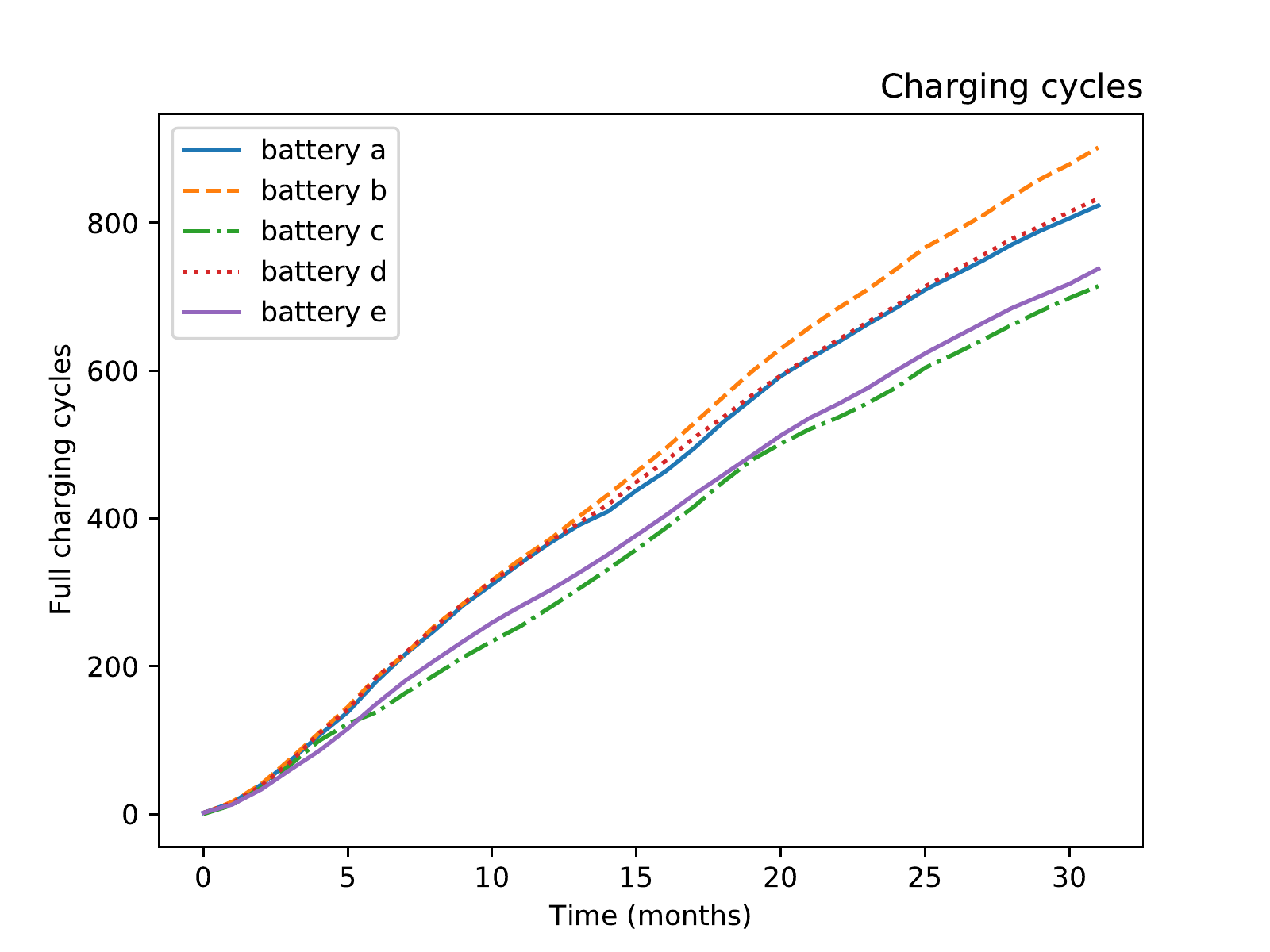}
\caption{The equivalent cycles for the batteries a, b, c, d and e}
\label{fig:cyc5}
\end{figure*}

Extrapolating from the equivalent cycle count graph, if the usage behavior remained unchanged, the model estimated the truck’s battery pack to complete around 3000 cycles in a 10-year service period (Figure \ref{fig:cyc5}). The estimated cycle life corresponds to the published cycle life for the commercial Nickel-Cobalt-Manganese (NMC) cells \cite{jal15cyc}, and support the findings in our previous study \cite{huo20dyn}.

\section{Discussion} \label{discussion}

In this paper, we have demonstrated the applicability of GB method for model development for battery state of health (SoH) predictions under circumstances when there is little prior information available about the batteries. It demonstrates how the supervised-learning-enabled SoH prognosis can effectively exploit the data from multiple cells in lithium-ion batteries from 45 EV forklifts and significantly improve the forecasting performance compared to previous models. 

In the model development work we could verify the results from the previous study with new data; e.g., the cycle life-time produced values that matched well with the values published by the cell manufacturers. This indicated that we were fairly successful in the development of the basic features and extracting their values. Furthermore, the developed GB model predicts the SoH well with the loss function value of MAE\textsubscript{SoH} 0.18 and with goodness of the fit, R$^2$ score 1 over a period of month.

Furthermore, we have shown that for a set of batteries, the Wilcoxon test yielded that the set of batteries come from the same distribution; this and the results using a common prediction model showed some promising loss function (MAE) and goodness of fit results for the set of batteries. However, it may well be that the different operating environments for the batteries result in SoH patterns that are not reliably enough captured by our GB model. Moreover, in the field of batteries, non-availability of more data sets with relatively long timeseries has is a known issue \cite{har17hyb}. This was also the case with our unique set of data as the timeseries were relatively short (32 months), which in turn indicates a drawback in our models. For example, the battery timeseries may show seasonality in the long run that we were unable to capture. 

With the GB mode, utilising the novel walk-forward algorithm for battery (a) over a period of three months, we yielded the MAE\textsubscript{SoH} loss function value 0.21 including the 95\% point-wise prediction confidence intervals. 
As for the environmental factors, the battery aging is known to be non-linear process, and the SoH deteriorates rapidly towards the end-of-life. For this reason, we introduced intrinsic frequency method to detect changes in the target SoH behavior for defining the point when the underlying GB model needs changes; the exact process how to implement this into the model needs further introspect. More data is needed to model end-of-lifetime behavior for the batteries, especially when taking into account that our model covered 3-year life-span out of around 10 years of expected life-time for the lithium-ion batteries for the EVs and for the 2nd life use. Although the model was developed with the data from EVs, it can be applied to the lithium-ion batteries that do not meet any longer the requirements of an EV application. It can still be used for the less-demanding grid-connected energy storage applications such as the battery energy storage system (BESS) for the sustainable energy management \cite{hefnawy2017combined}.

\section{Conclusions \& Future Work} \label{conclusion}

In this paper, we have demonstrated the applicability of the gradient boosting model for predicting battery state of health (SoH) timeseries under circumstances when there is little prior information available about the batteries. It demonstrates how supervised-learning-framed SoH prognosis can effectively exploit data from multiple cells in lithium-ion batteries from 45 EV forklifts to significantly improve the forecasting performance. The GB model predicts SoH well with the loss function value of MAE\textsubscript{SoH} = 0.21 and with goodness of the fit, R$^2$ score 1 over a period of three months, which is a reasonable time horizon in the context of preventive maintenance. Furthermore, we validated the model and rectified the symptoms of the overfit by utilising the novel  walk-forward algorithm; we yielded SoH predictions and the 95\% point-wise confidence intervals for the predictions. Moreover, the Huang-Hilbert transformation of the data provides some means to analyse if the SoH starts to deteriorate,  or,  to change less frequently or more frequently than before, which may indicate point-of-time to change the model; the transformation needs to be performed dynamically as the length of the data series changes. 

The future work could advance to extracting some user behavior patterns from the data as the number of the defined charging pulse timesteps (1000---2000) for each battery pack establish a basis for this kind of study. Also, an extension to this paper could be the further verification and development of the model for the battery SoH predictions with longer timeseries data, for example with the help of a relevant simulation, and further develop the model to make more robust predictions.

\section{Acknowledgments}

This work has been supported by the European Commission through the H2020 project FINEST TWINS (grant no. 856602), by the Academy of Finland through the project grants  296096 and 322742, by EUREKA Eurostars 2 (grant E! 12005 AIES) and by Business Finland. The authors appreciate the invaluable support of Professor Tanja Kallio and Mr. Panu Sainio from Aalto University in helping to evaluate the data and develop the research direction.

\bibliography{bibLiionXGB.bib}
\bibliographystyle{elsarticle-num}

\end{document}